\documentclass[sigconf]{acmart}
\usepackage{enumitem}  
\usepackage{color}
\usepackage{flushend}
\usepackage{graphicx}
\usepackage{booktabs}
\usepackage[htt]{hyphenat}
\usepackage[]{xcolor}
\usepackage{tabularx}
\usepackage{booktabs}
\usepackage{multirow}
\usepackage{xspace}
\usepackage{acronym}
\usepackage{arydshln}

\acrodef{CIS}{Conversational Information Seeking}
\acrodef{CS}{Conversational Search}
\acrodef{PTKB}{Personal Text Knowledge Base}
\acrodef{TREC}{TExt Retrieval Conference}
\acrodef{iKAT}{Interactive Knowledge Assistance Track}
\acrodef{CAsT}{Conversational Assistance Track}
\acrodef{NIST}{National Institute of Standards and Technology}
\acrodef{LLM}{Large Language Model}
\acrodef{NLU}{Natural Language Understanding}

\acrodef{IR}{Information Retrieval}
\acrodef{NLP}{Natural Language Processing}

\allowdisplaybreaks

\AtBeginDocument{%
  }

\copyrightyear{2025}
\acmYear{2025}
\setcopyright{acmlicensed}\acmConference[WWW Companion '25]{Companion Proceedings of the ACM Web Conference 2025}{April 28-May 2, 2025}{Sydney, NSW, Australia}
\acmBooktitle{Companion Proceedings of the ACM Web Conference 2025 (WWW Companion '25), April 28-May 2, 2025, Sydney, NSW, Australia}
\acmDOI{10.1145/3701716.3715869}
\acmISBN{979-8-4007-1331-6/2025/04}

\begin{document}

\title{Query Understanding in LLM-based Conversational \\ Information Seeking}


\author{Yifei Yuan}
\affiliation{%
  \institution{University of Copenhagen}
  \country{Denmark}
  }
\email{yiya@di.ku.dk}

\author{Zahra Abbasiantaeb}
\affiliation{%
  \institution{University of Amsterdam}
  \country{The Netherlands}
  }
\email{z.abbasiantaeb@uva.nl}

\author{Yang Deng}
\affiliation{%
  \institution{Singapore Management University}
  \country{Singapore}
  }
\email{ydeng@smu.edu.sg}

\author{Mohammad Aliannejadi}
\affiliation{%
  \institution{University of Amsterdam}
  \country{The Netherlands}
  }
\email{m.aliannejadi@uva.nl}

\renewcommand{\shortauthors}{Yuan
et al.}
\newcommand{\yifei}[1]{[\textcolor{orange}{#1}]}

\begin{abstract}
Query understanding in \ac{CIS} involves accurately interpreting user intent through context-aware interactions. This includes resolving ambiguities, refining queries, and adapting to evolving information needs. \acp{LLM} enhance this process by interpreting nuanced language and adapting dynamically, improving the relevance and precision of search results in real-time. In this tutorial, we explore advanced techniques to enhance query understanding in \ac{LLM}-based \ac{CIS} systems. We delve into \ac{LLM}-driven methods for developing robust evaluation metrics to assess query understanding quality in multi-turn interactions, strategies for building more interactive systems, and applications like proactive query management and query reformulation. 
We also discuss key challenges in integrating \acp{LLM} for query understanding in conversational search systems and outline future research directions. 
Our goal is to deepen the audience's understanding of \ac{LLM}-based conversational query understanding and inspire discussions to drive ongoing advancements in this field.

\end{abstract}

\maketitle

\section{Introduction}
Query understanding refers to a system’s ability to accurately interpret a user's intent, even when expressed through incomplete, vague, or ambiguous queries~\cite{Dai2019DeeperTU}. This becomes especially critical in \ac{CIS}, where users tend to express their needs in less direct or structured ways, compared to ad-hoc retrieval. In typical \ac{CIS} scenarios, users often begin with vague or imprecise queries and progressively refine their queries, ask follow-up questions, or even shift contexts mid-conversation~\cite{radlinski2017theoretical}. These dynamics pose significant challenges for the system, as it must engage in a dialogue to clarify and refine the user's intent, ensuring that the responses are both accurate and relevant throughout the interaction. 

In recent years, \acp{LLM}, such as GPT-4, have demonstrated remarkable capabilities in tasks far beyond \ac{NLU}, revolutionizing the way users interact with \ac{IR} systems. These models excel at handling natural language queries with exceptional accuracy and contextual awareness, greatly enhancing the overall user experience in retrieving relevant information. In the context of \ac{CIS} systems, \acp{LLM} significantly impact query understanding in several key areas, such as conversational context understanding~\cite{Mao2023LargeLM}, query clarification~\cite{Deng2023PromptingAE}, user simulation~\cite{Owoicho2023ExploitingSU}, query reformulation~\cite{Mo2023ConvGQRGQ}.


Despite the advancements \acp{LLM} have brought, several open challenges remain: (i) developing robust evaluation metrics, as it remains challenging to establish effective ways to measure how well a system understands and addresses user intent across dynamic, multi-turn conversations~\cite{Sekulic2022EvaluatingMC}; (ii) Improving conversational interaction by making exchanges smoother and more natural~\cite{Lei2020EstimationActionReflectionTD}; (iii) increasing user proactivity, encouraging users to take a more active role in refining and clarifying their searches~\cite{Deng2023PromptingAE}; and (iv) handling ambiguity in user queries, as users frequently submit vague or incomplete queries, requiring \acp{LLM} to strike a balance between generating appropriate responses and requesting clarifications~\cite{Keyvan2022HowTA}. 
This tutorial addresses these challenges by exploring advanced techniques to improve query understanding in \ac{LLM}-based \ac{CIS}.



\subsection{Query Understanding Evaluation}
Query understanding refers to the process where a system interprets the intent and context of a user’s query to deliver more accurate and relevant search results~\cite{dalton2020trec}. Evaluating query understanding involves assessing how accurately a system interprets and responds to user queries in alignment with their intent~\cite{Wang2015QueryUT}. We discuss two main sets of works:

\begin{itemize}[leftmargin=*,nosep]
   
    \item \textbf{End-to-end evaluation} utilizes human-judged benchmarks to assess the relevance of query-passage pairs. Among these benchmarks, QReCC~\cite{anantha2021qrecc} and TopioCQA~\cite{adlakha2022topiocqa} are two large-scale open-domain conversational question-answering datasets.
    TREC CAsT 19-22~\cite{owoicho2022trec,dalton2020trec} and TREC iKAT 23~\cite{Aliannejadi2023ikat} benchmarks feature complex, knowledge-intensive conversations. 
 \item  \textbf{\ac{LLM}-based relevance assessment}
leverages \acp{LLM} to evaluate the relevance of retrieved information to a user’s query~\cite{abbasiantaeb2024uselargelanguagemodels,meng2024query,khramtsova2024leveraging}. However, using \acp{LLM} comes with challenges such as non-reproducibility, unpredictable outputs, and potential data leakage between fine-tuning and inference stages~\cite{pradeep2023rankzephyr}. 

    
\end{itemize}

\subsection{\ac{LLM}-based Conversational Interaction}  
\ac{LLM}-based conversational interactions improve query understanding through dynamic, back-and-forth exchanges that clarify user intent and enhance search precision. Unlike static searches, conversational \acp{LLM} capture nuances and progressively build context. We focus on two key aspects:
\begin{itemize}[leftmargin=*,nosep]
    \item \textbf{\ac{LLM}-based user simulation.} Simulating diverse user behaviors, intents, and query patterns helps \acp{LLM} learn to anticipate real-world conversational scenarios, preparing them to handle complex queries and varied user needs effectively~\cite{Wang2023UserBS,Wang2023AnII}. This approach has become essential in evaluating systems across domains such as information-seeking dialogues~\cite{Ren2024BASESLW,Sekuli2024AnalysingUI}, conversational question-answering~\cite{Abbasiantaeb2023LetTL}, and task-oriented dialogues~\cite{Sekulic2024ReliableLU}.
    \item \textbf{Multimodal conversational interactions.} ~ Integrating beyond-text content (e.g., images, audio) into conversations enables multimodal interactions, allowing \acp{LLM} to interpret and respond across diverse media types. This capability has expanded applications in areas like e-commerce, healthcare, and spatial analysis, enhancing tasks such as \ac{LLM}-powered multimodal fashion search~\cite{Goenka2022FashionVLPVL,Yuan2021ConversationalFI}, medical image retrieval~\cite{Tiwari2023ExperienceAE}, and beyond.
    
\end{itemize}

\subsection{\ac{LLM}-based Proactive Query Management}

Conventional \ac{CIS} systems passively respond to user queries.
For example, current CIS systems may refuse to answer or provide low-quality answers when encountering unanswerable user queries.
Here, we will introduce recent advances in developing \ac{LLM}-based proactive CIS systems that can further provide useful information to unanswerable queries, or clarify the uncertainty of the query for more efficient and precise information seeking. In particular, we will cover the following:

\begin{itemize}[leftmargin=*,nosep]
    \item \textbf{Unanswerable query mitigation.} Typically, the system handles unanswerable queries passively by responding with \texttt{No Answer}~\cite{quac} if there is no direct information that matches the query. This undesired result will downgrade the user experience when interacting with the CIS systems. Researchers investigate various proactive behaviors to mitigate this issue, including providing relevant information that can partially satisfy the user’s information needs~\cite{inscit} or explanations on why the query is unanswerable~\cite{deng2024gotcha}, and suggesting other useful queries~\cite{www20-query-suggest,www24-query-suggest}.  
    
    \item \textbf{Uncertain query clarification.} Asking clarifying questions allows the users to further clarify their queries in case the model is uncertain about their intent~\cite{sigir19-clarify,Aliannejadi2021BuildingAE,Zhang2024CLAMBERAB}. Recent studies develop various training paradigms to teach \acp{LLM} to ask clarifying questions, such as in-context learning~\cite{Deng2023PromptingAE}, self-learning~\cite{colm24-star-gate}, reinforcement learning~\cite{acl24-style}, and contrastive learning~\cite{Chen2024ACT}, as well as in multimodal scenarios~\cite{Yuan2024AskingMC}. 
    
    \item \textbf{Balancing user and system initiatives.} Taking the conversation initiative by the system introduces a great risk of harming user experience~\cite{zou2023asking}, while not necessarily leading to improved retrieval~\cite{krasakis2020analysing}. Therefore, it is of utmost importance to learn ``when'' to take the initiative in a conversation. Recent work argues that \acp{LLM} are not capable of effective planning for taking system-initiative actions~\cite{shaikh2024grounding,aliannejadi2024interactions}. A solution is to predict for the system when to take the initiative in a conversation~\cite{Wadhwa2021towards,meng2023system} while simulating user--system interactions~\cite{aliannejadi2021analysing} is used to understand the dynamics of system initiative better.
\end{itemize}  

\subsection{\ac{LLM}-based Query Enhancement}
Query enhancement is the process of modifying a user's original query to improve retrieval performance and enhance the accuracy of search performance. 
By reformulating the query, systems can better match the user's intent, leading to more relevant and precise results. We cover several interactive query enhancement techniques where \acp{LLM} interpret and refine queries to capture deeper semantic nuances and better understand user intent, namely: 

\begin{itemize}[leftmargin=*,nosep]
    \item \textbf{Resolving ambiguity in queries.} Query ambiguity has been investigated in many studies from various aspects such as automatic ambiguous query detection and introducing taxonomy of queries~\cite{Keyvan2022HowTA}. To resolve ambiguous queries, \ac{LLM}-based techniques such as query expansion~\cite{Jagerman2023QueryEB,Wang2023Query2docQE}, query refinement~\cite{Dhole2023AnIQ}, and follow-up question suggestion~\cite{Baek2023KnowledgeAugmentedLL} have proven effective. These approaches help clarify intent and guide users toward more precise search results.
    
    \item \textbf{Conversational query rewriting} is the process of rephrasing or modifying a user’s query within a conversational context to improve retrieval accuracy and relevance~\cite{vakulenko2021question,raffel2020exploring}. \acp{LLM} enhance query rewrite performance in several ways: (i) handling low-resource (few-shot or zero-shot) scenarios~\cite{ye-etal-2023-enhancing,Mao2023LargeLM,yu2020few}; (ii) incorporating multimodal contents to improve the rewrites~\cite{yuan2022mcqueen}; and (iii) generating LLM-based answer for better retrieval~\cite{abbasiantaeb2024generate}.
    
\end{itemize}

\subsection{Open Challenges and Beyond}
In the final part, we will explore key open challenges in integrating LLMs for query understanding within conversational search systems and outline research directions for future investigation.
\begin{itemize}[leftmargin=*]
    \item \textbf{Multilingual and cross-cultural query understanding.} While LLMs perform reasonably well in understanding English queries, challenges persist in handling queries across diverse languages and cultural contexts. Expanding LLM capabilities to better support multilingual and culturally nuanced queries is essential for fostering more inclusive and accurate search experiences.
   \item \textbf{Real-time adaptation to evolving user intent.} Developing models that can dynamically adjust search strategies in response to evolving user intent throughout a conversation remains a significant challenge for \ac{CIS} systems~\cite{Sun2017CollaborativeIP}. Under this context, instructing LLMs to accurately detect and adapt to shifts in user intent remains an important future direction.
 \end{itemize}

\section{Relevance to the Conference}
This tutorial is highly relevant to The Web Conference as it addresses critical challenges in advancing conversational AI and \ac{IR} systems --- two fields that are integral to the future of web interactions. By exploring the latest techniques to improve LLM-based conversational systems, this tutorial aligns with the conference's mission to drive innovations in web technologies and enhance user experiences on the web. 

Related tutorials in recent years include: (i) \textit{Conversational Information Seeking: Theory and Application (SIGIR22)}~\cite{Dalton2022ConversationalIS}; (ii) \textit{Proactive Conversational Agents in the Post-ChatGPT World (SIGIR23)}~\cite{Liao2023ProactiveCA}; (iii) \textit{Large Language Model Powered Agents in the Web (WWW24)}~\cite{Deng2024LargeLM}; (iv) \textit{Tutorial on User Simulation for Evaluating Information Access Systems on the Web (WWW24)}~\cite{Balog2024TutorialOU}. However, these tutorials mainly introduce applications of conversational IR and agent-based interactions. Our tutorial mainly focuses on enhancing query understanding within \ac{LLM}-based conversational IR systems and beyond. 
    
\section{Detailed Schedule}
This tutorial will be a \textbf{lecture-style} tutorial focusing on the latest advancements in query understanding based on \ac{LLM}-powered \ac{CIS} systems. The outline of this tutorial is summarized as follows: 
\begin{itemize}[leftmargin=*,nosep]
    \item \textbf{Introduction} (20 min): ad-hoc search; preliminary of query understanding; adapting \acp{LLM} in query understanding.
    \item \textbf{Part I: conversational query understanding evaluation} (30 min): 
    end-to-end evaluation; 
    \ac{LLM}-based relevance assessment.
    \item \textbf{Part II: \ac{LLM}-based conversational 
    interaction} (30 min): 
    \ac{LLM}-based user simulation; 
    multimodal conversational interaction. \looseness=-1
    \item \textbf{Part III: \ac{LLM}-based proactive query management} (40 min): 
    unanswerable query mitigation; 
    ambiguous query clarification;
    balancing user and system initiatives. \looseness=-1
    \item \textbf{Part IV: \ac{LLM}-based query enhancement} (30 min): 
    resolving ambiguity in queries; 
    conversational query rewrite techniques.
    
    \item \textbf{Summary and outlook} (30 min): open challenges and beyond.
    
\end{itemize}

\section{Target Audience and Materials}
This tutorial is designed for researchers, students, and anyone interested in \ac{LLM}-based conversational search, query understanding, \ac{IR}, data mining, and NLP. The target audience includes \textbf{NLP and IR Researchers}: those exploring how \acp{LLM} enhance conversational query understanding and search. \textbf{Conversational AI Practitioners}: professionals developing AI-driven chatbots, virtual assistants, and support systems. \textbf{Graduate Students and Academics}: early-career researchers and students looking to apply \acp{LLM} in \ac{CIS}.


We will create a website for all relevant materials: (i) \textbf{a presentation slide} covering the background, technique, and future directions discussed in the tutorial; (ii) \textbf{a video teaser} for public promotion \footnote{\url{https://drive.google.com/file/d/1UMki52eKDXMnph3ifl9bM0lhV55wFYc9/view?usp=sharing}}; and (iii) \textbf{annotated reference} to enable further study. 
\section{BIOGRAPHY OF PRESENTERS}
\textbf{Yifei Yuan} is a Postdoctoral Research Fellow at the University of Copenhagen. She received her Ph.D.\ degree from the Chinese University of Hong Kong in 2023 and B.Eng. degree from Harbin Institute of Technology in 2019. Her research interests lie in \ac{NLP} and \ac{IR}, especially for conversational interactive search systems and image-text-based multimodal learning. She has published more than 15 papers on relevant topics at top conferences in \ac{NLP} and Data Mining. She has been serving as a reviewer or program committee member of mainstream machine learning venues such as ICLR, ACL, SIGIR, and WWW. 


\noindent\textbf{Zahra Abbasiantaeb} is a second-year Ph.D.\ student at the Information Retrieval Lab (IRLab), University of Amsterdam (UvA). She received her master's degree from Amirkabir University (Tehran), on Artificial Intelligence in 2021. Her research interests lie in \ac{IR} and \ac{CIS} systems.  
She has published several papers at top conferences including SIGIR and WSDM. She is co-organizing the interactive Knowledge Assistant Track (iKAT) at the Text REtrieval Conference (TREC), aiming to advance the development of personalized conversational search systems. 


\noindent\textbf{Yang Deng} is an Assistant Professor at Singapore Management University.  
His research lies in \ac{NLP} and \ac{IR}, especially for conversational and interactive systems. 
He has published over 50 papers on relevant topics at top venues such as WWW, SIGIR, ACL, EMNLP, and ICLR, and serves as Area Chair for ACL, EMNLP, and NAACL. 
He has rich experience in organizing tutorials at top conferences, including WWW 2024, SIGIR 2024, and ACL 2023. 



\noindent\textbf{Mohammad Aliannejadi} is an Assistant Professor at IRLab, University of Amsterdam. 
His research interests include conversational information access, recommender systems, and \ac{LLM}-based data augmentation and evaluation.
Mohammad has co-organized various evaluation campaigns such as TREC CAsT, TREC iKAT, CLEF Touché, ConvAI3, and IGLU, focusing on different aspects of user interaction with conversational agents. 
Moreover, Mohammad has held multiple tutorials and lectures on \ac{CIS}, such as ECIR, SIGIR-AP, WSDM, CHIIR, SIKS, and ASIRF.


\bibliographystyle{ACM-Reference-Format}
\bibliography{sample-base}

\end{document}